\documentclass{article}

\usepackage{arxiv}
\usepackage{dsfont,amsmath}
\usepackage[utf8]{inputenc} 
\usepackage[T1]{fontenc}    
\usepackage{hyperref}       
\usepackage{url}            
\usepackage{booktabs}       
\usepackage{amsfonts}       
\usepackage{nicefrac}       
\usepackage{microtype}      
\usepackage{cleveref}       
\usepackage{lipsum}         
\usepackage{graphicx}
\usepackage{natbib}
\usepackage{doi}

\usepackage[onehalfspacing]{setspace}
\newcommand{\unmarkedfootnote}[1]{%
  \renewcommand{\thefootnote}{}
  \footnotetext{#1}
  \renewcommand{\thefootnote}{\arabic{footnote}}
}

\title{A graph generation pipeline for critical infrastructures based on heuristics, images and depth data*}

\date{March 6, 2026}

\newif\ifuniqueAffiliation
\uniqueAffiliationtrue

\author{ \href{https://orcid.org/0000-0001-9838-0862}{\includegraphics[scale=0.06]{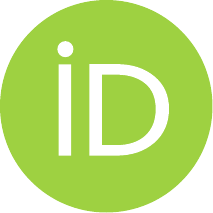}\hspace{1mm}Mike Diessner} \\
	German Aerospace Center (DLR)\\
	Institute for the Protection of Terrestrial Infrastructure \\
	Sankt Augustin, Germany \\
	\texttt{mike.diessner@dlr.de} \\
	\And
	Yannick Tarant \\
	German Aerospace Center (DLR)\\
	Institute for the Protection of Terrestrial Infrastructure \\
	Sankt Augustin, Germany \\
	\texttt{yannick.tarant@dlr.de} \\
}

\renewcommand{\headeright}{}
\renewcommand{\undertitle}{}
\renewcommand{\shorttitle}{Graph generation for critical infrastructure}

\hypersetup{
pdftitle={A graph generation pipeline for critical infrastructures based on heuristics, images and depth data},
pdfsubject={},
pdfauthor={Mike Diessner, Yannick Tarant},
pdfkeywords={Graph generation, critical infrastructure, photogrammetry, depth data, image data, object detection, instance segmentation, digital twin},
}

\begin{document}
\maketitle

\begin{abstract}
Virtual representations of physical critical infrastructures, such as water or energy plants, are used for simulations and digital twins to ensure resilience and continuity of their services. These models usually require 3D point clouds from laser scanners that are expensive to acquire and require specialist knowledge to use. In this article, we present a prototypical graph generation pipeline based on photogrammetry. The pipeline detects relevant objects and predicts their relation using RGB images and depth data generated by a stereo camera. This more cost-effective approach uses deep learning for object detection and instance segmentation of the objects, and employs user-defined heuristics or rules to infer their relations. Results of two hydraulic systems show that this strategy can produce graphs close to the ground truth. While this study focuses on hydraulic systems, the general process can be used to tailor the method to other types of infrastructures and applications. The user-defined rules create transparency qualifying the pipeline to be used in the high stakes decision-making that is required for critical infrastructures.\unmarkedfootnote{* Journal reference: Mike Diessner and Yannick E Tarant. A graph generation pipeline for critical infrastructures based on heuristics, images and depth data. \textit{Frontiers in Signal Processing}, 6:1761293, 2026. doi:\href{https://doi.org/10.3389/frsip.2026.1761293}{10.3389/frsip.2026.1761293}}
\end{abstract}



\section{Introduction}\label{sec:introduction}

Critical infrastructures are assets, organizations or facilities that are integral to the survival and functioning of a nation and its society. Disturbances in their services can have major adverse impacts, such as supply shortages, increased risks to public health and national security, or disruptions of financial services and the broader economy. Critical infrastructures can be physical or virtual assets and appear in sectors such as energy, healthcare, food, water, transportation, communication, nuclear systems and more \citep{osei2021, alcaraz2015, yusta2011}.

Methods such as simulations and digital twins are used to monitor, maintain and optimize infrastructures, and to prepare and train for worst-case scenarios by simulating extreme and often rare events. Thus, these methods have the potential to increase resilience, security, performance and continuity of infrastructures and are an invaluable tool to ensure national security and wellbeing \citep{lampropoulos2024, sousa2021}. In many cases, a digital twin of a physical asset, such as a nuclear facility or a water treatment plant, requires a virtual copy of the objects (e.g., pipes, pumps and valves) and their relation to each other to be used for simulations or training in virtual reality. The process of generating these models requires a large amount of manual work and is thus expensive and time-consuming \citep{franke2023}. Most existing methods use laser scanners to gather 3D point clouds of the facilities that are then processed and analyzed. However, these scanners are expensive, creating demand for methods using more affordable and accessible equipment, such as stereo cameras.

The main objective of this article is the development of an algorithm that uses cost-effective photogrammetry \citep{Moon2019}, i.e., RGB images, depth data and camera positions, to automatically detect relevant objects within a building and predict the relations between these objects. The algorithm should yield a graph network, where nodes represent objects sand edges represent physical connections between objects. As the main use of this pipeline is for high stakes decision making for critical infrastructures, the algorithm design will focus on explainability and interpretability besides accessibility and cost-effectiveness, where possible. While we were able to validate the pipeline on synthetic scenes that mimic the real world, it remains a prototype that requires validation on actual real-world scenes.

This article is structured as follows. Section~\ref{sec:relatedworks} places our proposed method into context by giving an overview of the related works. Section~\ref{sec:methods} details the developed graph generation pipeline and the data of the hydraulic systems used as the test environments for validation. Section~\ref{sec:results} presents the results of applying the proposed pipeline to two hydraulic systems. Section~\ref{sec:sensitivity} conducts a sensitivity and runtime analysis to investigate the robustness and practical feasibility of the pipeline. Lastly, Section~\ref{sec:discussion} outlines the strength and limitations of the pipeline and Section~\ref{sec:conclusion} draws a brief conclusion.


\section{Related works}\label{sec:relatedworks}

Most of the time, a virtual geometric representation of a building or a facility is the starting point of a digital twin. This representation is then enriched with attributes and information from individual objects, sensors and real-time data \citep{Jones2020, Grieves2014}.In the last decade, frameworks such as building information modelling (BIM) have been used increasingly in the planning and building of new facilities and can provide the required geometric and attribute data \citep{Ghaffarianhoseini2017, Borrmann2018, Lu2022}. While the adoption of building information modelling has grown rapidly in the last decade, it was likely not used for most older existing buildings. In these cases, the first step in building a digital twin is the collection of geometric data, the identification of relevant objects and the extraction of key attributes \citep{Borkowski2023}.

A popular method for collecting geometric data is the use of 3D laser scans. Laser scanners create a three-dimensional point cloud with a high resolution providing a detailed virtual representation of reality. However, these scanners are expensive and can cost many tens to hundreds of thousands of dollars. They are also heavy and require expensive software for processing the acquired data. Furthermore, specialized training is needed for operation and they struggle to represent reflective surfaces correctly \citep{Moon2019}. An alternative approach is photogrammetry that aims to generate precise 3D models from photographs and usually involve the computation of depth data. Multi-view stereo, for example, uses multiple perspective of a stereo camera or a RGB and a thermal camera to compute the depth information and a 3D scene \citep{Goesele2006, Vidas2013}, while Structure-from-Motion uses a series of 2D images to estimate the depth data and camera positions \citep{Schonberger2016}. Overall, systems used for photogrammetry cannot achieve the same level of detail and resolution as 3D laser scanners but are cheaper, more accessible and more mobile. 

3D laser scanning and photogrammetry provide two different outputs: point clouds and images. Both present distinct strength and shortcomings as inputs for object detection. Albeit that point clouds have a high resolution and are very accurate they are also sparse and the point densities can be highly variable leading to high computational requirements and costs \citep{Zhou2018}. Furthermore, there are a limited number of labeled data sets for point clouds available as labelling in three-dimensional space is challenging \citep{Zimmer2022}. This also makes collecting and labelling custom data sets for training difficult and time-intensive. Despite of this, point clouds were used in object detection for hydraulic systems in various articles \citep{Qiu2014, Kawashima2014, Cheng2020, Alex2025}, while the use of photogrammetry is limited \citep{Hart2023, Zhao2025}. However, as object detection on images is computationally cheaper and more training data is available, easier to collect and label, it lends itself for object detection as part of a graph generation pipeline that aims to be widely accessible, applicable to different use cases, and time- and cost-effective. Furthermore, models for object detection on images are generally more researched and thus more mature than their counterparts using point clouds. This is also reflected in the many models that are available, many of which even offer good out-of-the-box performance \citep{redmon2016, kirillov2023, he2017, yuan2021}. In practice, we found that object detection on the many different perspectives of the images was reliable and results from different view points had the advantage of enabling validation of detections across images. The latter was especially valuable for featureless objects and instances in which lights and reflections made object detection challenging as there were multiple opportunities to detect the same object. This decreased the possibility of erroneous detections. Thus, we identified photogrammetry in combination with object detection on images as the superior methods for the pipeline presented in this article. While the presented pipeline is closely linked to multi-view stereo it is not just a geometric 3D representation of a scene but also aims to encode relational information between objects of the scene in a graph. 

The prediction of a relational graph of the objects contained an image, also known as scene graph generation, through graph neural networks has seen increased interest by the deep learning community in recent years \citep{shit2022, yang2018, li2024, cong2023}. These approaches combine the detection of relevant objects in an image with the prediction of relations between the detected objects and encode both in a graph. Similar to other deep learning methods, these models are black boxes lacking explainability and interpretability \citep{Buhrmester2021, Sahin2025} and although there are efforts to remedy this in the form of explainable AI \citep{Xu2019, zhang2022, Dwivedi2023, peng2024}, it is questionable if these efforts suffice when it comes to high stakes decision-making \citep{Rudin2019}. Furthermore, scene graph generation networks require data and additional labeling and cannot easily be transferred to other use cases and scenes involving different objects without collecting new data. For example, many graph generation networks are based on the 3DSSG data set that contains 1,482 scene graphs with 48k objects and 544k relations \citep{wald2019, wald2020, lv2024, yeo2025}. Collecting and annotating a data set of this size for a specific use case is, in practice, at least impracticable if not infeasible. An alternative to these models is an approach using heuristics and rules based on which relations between objects are inferred. Letting the user define a custom set of rules gives them full control and enables them to tailor the method to a specific application or transfer it to another without the need of collecting additional training data. While there is some initial manual work involved in setting up the rules, it is likely negligible compared to the effort of collecting and labelling data for the graph neural networks. This article chooses the latter approach due to its flexibility and cost-effectiveness but mainly because the rules make the inner workings of the algorithm transparent and explainable which is essential for high stakes decision-making as it is required in critical infrastructure \citep{Rudin2019}. The process of defining these rules and their requirements are discussed in Section~\ref{sec:graph}.


\section{Materials and Methods}\label{sec:methods}
 
\subsection{Data}\label{sec:images}

\begin{figure}[!ht]
    \begin{center}
        \includegraphics{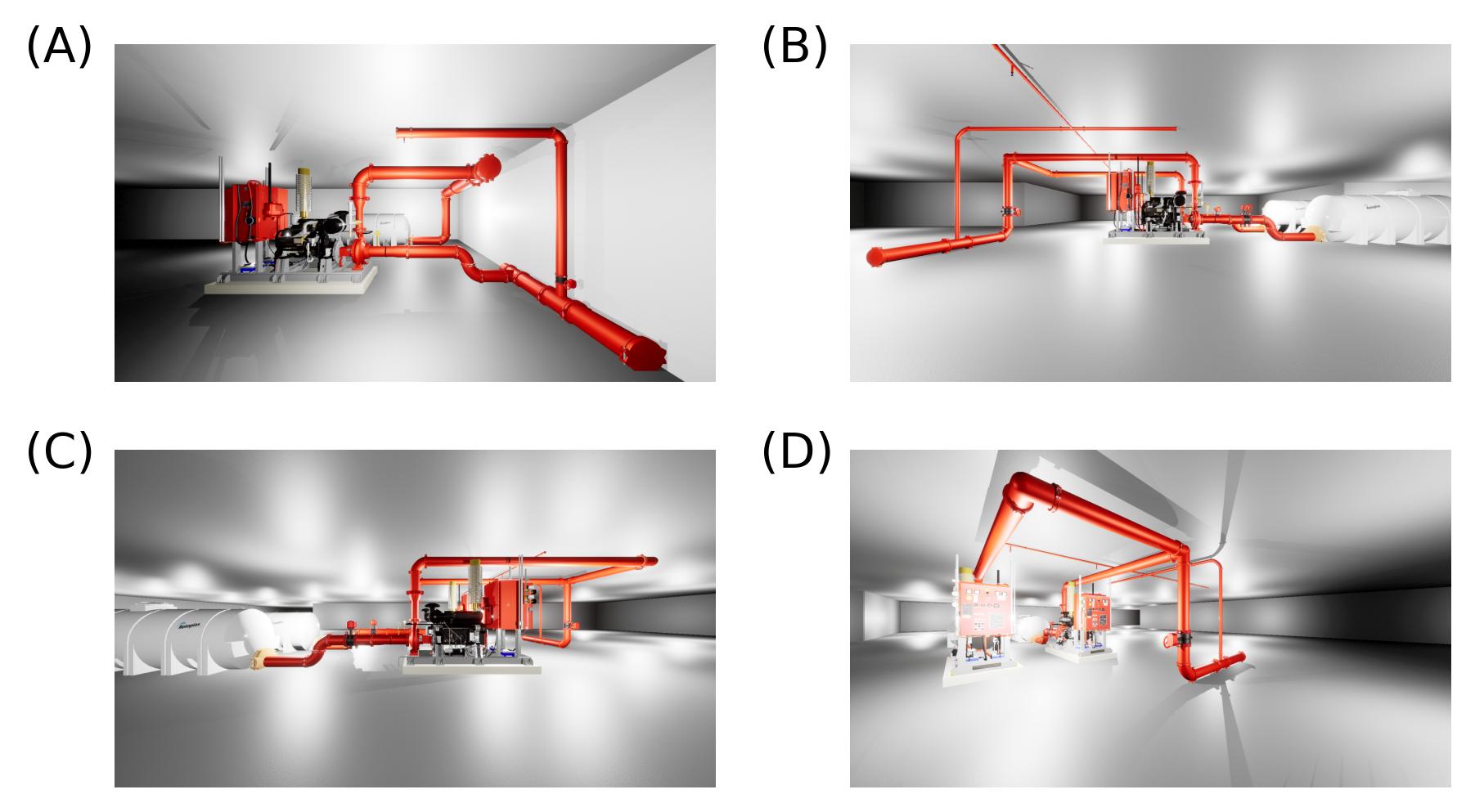}
    \end{center}
    \caption{Synthetic hydraulic systems. (\textbf{A}) shows system 1 consisting of pipes, one pump, one tank, one valve and one sprinkler. (\textbf{B}), (\textbf{C}) and (\textbf{D}) show system 2 consisting of pipes, two pumps, two tanks, three valves and four sprinklers.}\label{fig:1}
\end{figure}

The nature of critical infrastructures prohibits the use of images and data collected at the actual real-world sites due to security concerns. Thus, we implement virtual representations mimicking the most important characteristics of their real-world counterparts in Unreal Game Engine 5 \citep{epicgames2022a}. We chose the Unreal Engine as it allows for rapid generation of realistically rendered scenes and contains tools to provide global dynamic illumination \citep{epicgames2022c}, rendering of meshes with arbitrary resolution \citep{epicgames2022b} and scripted capturing of realistic images in combination with segmentation masks. We use Colosseum \citep{colosseum2022}, a fork of AirSim \citep{airsim2017}, to simulate cameras in the Unreal Engine allowing us to produce RGB images and depth data along with the intrinsic and extrinsic camera parameters, such as the position and orientation of the camera at the time of taking the images. While real-world testbeds are preferable, the realistic results of the Unreal Engine strike a good balance between accessibility and an accurate representation of the real world.

This paper focuses on the graph generation of hydraulic systems by investigating two test environments. Hydraulic system 1 consists of a pump, a tank, a valve, a sprinkler and pipes as shown in Figure~\ref{fig:1} (\textbf{A}). Hydraulic system 2 increases the complexity by adding another pump and tank, two valves and three sprinklers to the design. This system is shown from multiple perspectives in Figure~\ref{fig:1} (\textbf{B}), (\textbf{C}) and (\textbf{D}). Pipes, pumps, tanks, valves and sprinklers are the only objects in these test environments. It should be noted that the distinct color of the pipes compared to the floor, walls and ceiling is chosen for illustrative purposes only. The components of the pipeline, such as the models used for object detection, are not reliant on the color of the pipes as they were trained and tested on a wide range of different color and texture compositions. We use domain randomization to randomly generate different color and texture combinations for the pipe objects and use them to produce a diverse set of training images via the Unreal Engine for fine-tuning YOLOv8 as outlined in more detail in \cite{schreiber2024}.

Data for hydraulic system 1 comprises 16 RGB and depth image pairs with a resolution of 1920 $\times$ 1080 pixels and a field-of-view of 114°. For the more complex hydraulic system 2, we stock up the number of RGB and depth image pairs to 29 with identical resolution and field-of-view.

The training data for YOLOv8 consists of photos of hydraulic systems taken with a Canon EOS 7D SLR camera at a resolution of 6000 $\times$ 4000 pixels and synthetic images generated with Unreal Engine 5 as non-interlaced 8-bit RGBA images with a resolution of 1920 $\times$ 1006 pixels.

\subsection{Pipeline}\label{sec:pipeline}

\begin{figure}[!ht]
    \begin{center}
        \includegraphics{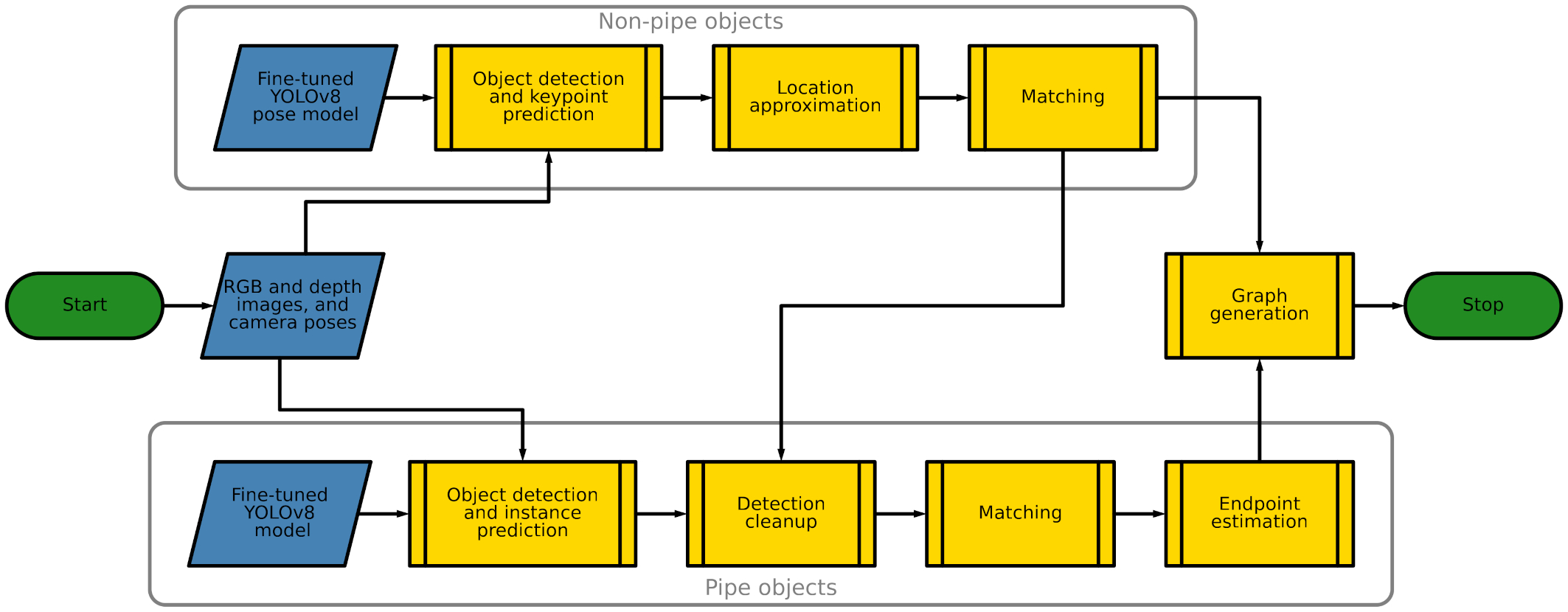}
    \end{center}
    \caption{Flowchart of the graph generation pipeline. Pipe and non-pipe objects are treated differently and are combined in the 'Graph generation' module.}\label{fig:2}
\end{figure}

The graph prediction pipeline is structured as shown in Figure~\ref{fig:2}. The pipeline differentiates between pipe and non-pipe objects and processes them with two distinct approaches. Both use RGB images, depth data and camera poses including the camera position and camera orientation of the individual images and a fine-tuned YOLOv8 model \citep{redmon2016}. While we chose YOLOv8 for object detection, the pipeline in this study does not require the use of YOLOv8. The detection model is solely an input of the pipeline and can be replaced with any preferred model. We decided to use YOLOv8 as it is an established model that does not require a large data set for fine-tuning the pre-trained model to a new detection task as it is the case for vision transformers \citep{dosovitskiy2021}. It also provides versions for instance segmentation and keypoint prediction that are used as described in the following sections.

\subsubsection{Non-pipe objects}\label{sec:nonpipe}

\begin{figure}[!ht]
    \begin{center}
        \includegraphics{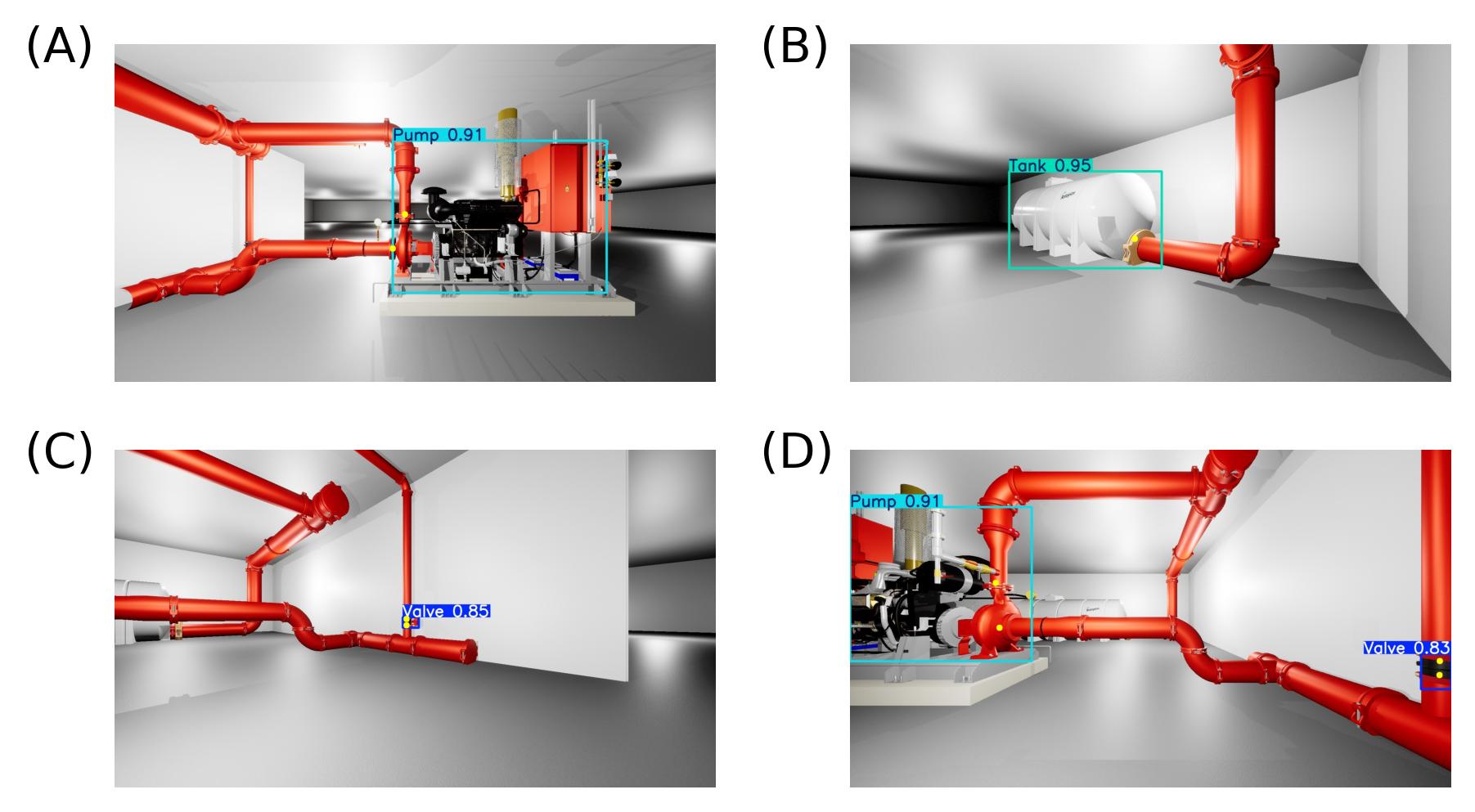}
    \end{center}
    \caption{Object detection of pumps, tanks and valves including one keypoint for tanks and two keypoints for pumps and valves. Bounding boxes indicate detected objects with class label and class confidence score. Keypoints are indicated by yellow dots. (\textbf{A}) shows a pump, (\textbf{B}) shows a tank, (\textbf{C}) shows a valve and (\textbf{D}) shows a different perspective of the pump and the valve.}\label{fig:3}
\end{figure}

Non-pipe objects are processed using a fine-tuned YOLOv8 pose model \citep{redmon2016} that detects the pumps, tanks and valves within the images and predicts one keypoint for tanks and two keypoints for pumps and valves. These keypoints represent the connection points of the tanks, valves and pumps with other objects, such as pipes. The largest YOLOv8 architecture pre-trained on the COCO data set \citep{lin2014} was chosen as a foundation. This model was fine-tuned using a data set consisting of 526 images, with 280 being real-world RGB images and the remaining 246 being synthetically created using Unreal Engine 5 \citep{epicgames2022a}. The training was conducted for 150 epochs using the Adam optimizer and a learning rate of 10\textsuperscript{-4}.  Figure~\ref{fig:3} shows predictions of the model for four images. Detections for an object consist of a bounding box with a class label and a class confidence score, and keypoints plotted in yellow. For example, Figure~\ref{fig:3} (\textbf{A}) shows a pump detected with a confidence of 91~\% and two keypoints indicating the connection points with the neighboring pipes. YOLOv8 uses Non-Maximum-Suppression (NMS) to discard redundant detections by only using bounding boxes with the highest confidence score and dismissing other overlapping bounding boxes with lower confidence scores.

The second module in the 'Non-pipe objects' branch of the pipeline aims to approximate the location of the objects within the bounding boxes. While it would be possible to use an instance segmentation method to get exact masks of the objects (similar to the segmentation used for the pipes in Section~\ref{sec:pipe}), exact masks are not required for the pipeline to achieve satisfying results. Hence, we opt for an approximation of object masks sparing us time-consuming instance labelling of the more intricate non-pipe objects. The main objective of the location approximation is to delete any background pixels within the predicted bounding boxes, leaving us with only pixels belonging to the object. Experimentation showed that we can achieve this objective by chaining three methods together: (i) We disregard any pixels with depth values larger than the median of all depth values within a bounding box. This eliminates the majority of the background. (ii) We refine the mask by projecting the remaining pixels to a point cloud down-sampled to one voxel per cm\textsuperscript{3} and removing any statistical outliers (voxels that deviate more than one standard deviation from a neighborhood consisting of the closest 25~\% of all other voxels). (iii) Finally, we employ Density-Based Spatial Clustering of Application with Noise (DBSCAN) to approximate the location of the object \citep{ester1996, schubert2017}. We setup DBSCAN such that points that are less than 2 cm apart are assigned to a cluster. Finally, the largest cluster is taken as the approximated location.

\begin{figure}[!ht]
    \begin{center}
        \includegraphics{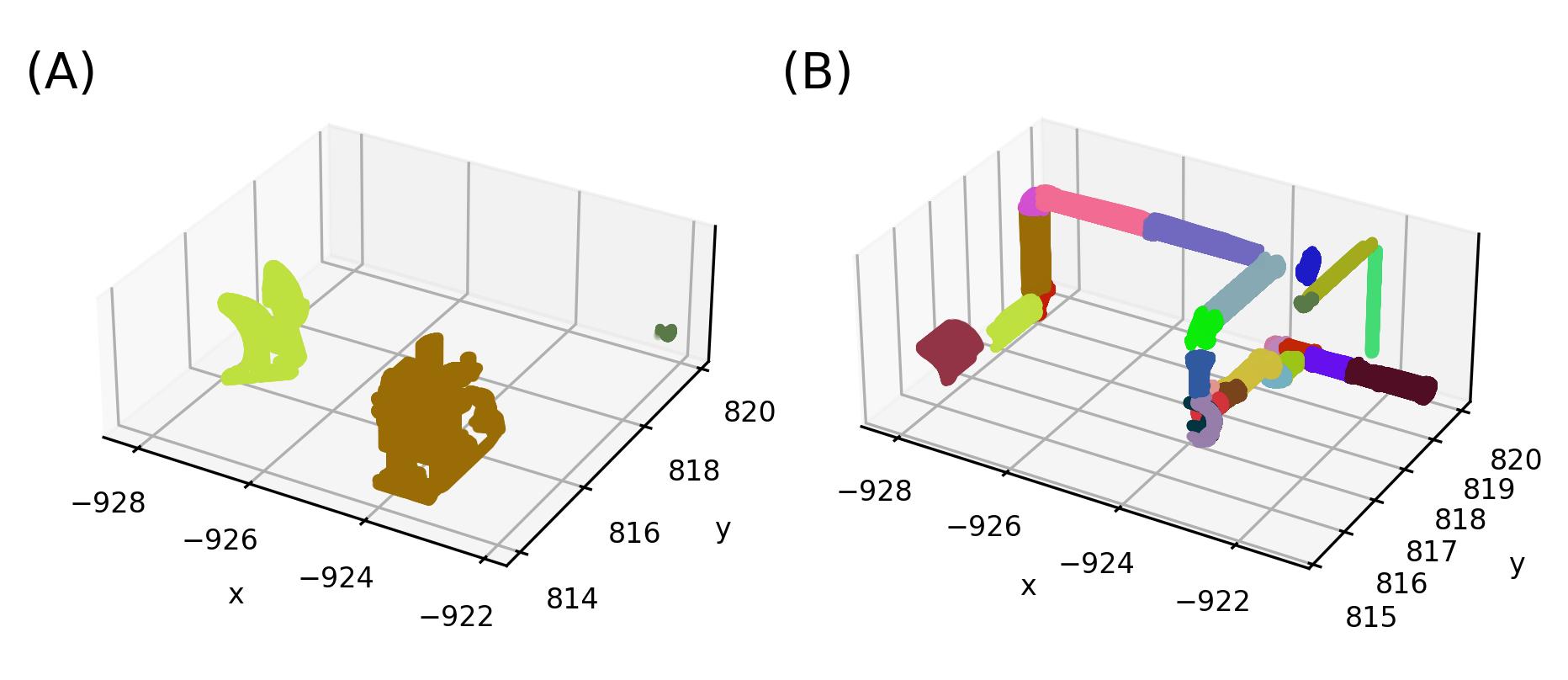}
    \end{center}
    \caption{Object matching across individual images. (\textbf{A}) shows the matched objects of a pump, a tank and a valve and (\textbf{B}) shows the matched pipe objects. Each color indicates a distinct object.}\label{fig:4}
\end{figure}

The last module in processing the non-pipe objects matches identical detections across images and combines information in one final object. In this step, the approximated locations of merged objects are combined and their endpoints (predicted via the keypoints) are averaged. Before averaging, possible false predictions are disregarded via statistical outlier removal of keypoints that deviate more than two standard deviations from all keypoints. Two objects from different images are merged if the following rule for the pairwise distances $\textbf{d}$ between all their voxels is true:

$$\frac{\sum_{i=1}^n \mathds{1}_{\{d_i < np\_max\_distance\}}}{n}  > np\_min\_percentage,$$

where $n$ is the number of elements of vector $\textbf{d}$, and $np\_max\_distance \in (0, \infty)$ and $np\_min\_percentage \in [0, 1]$ specify the maximal distance in meters between two object points to be considered a match and the minimal percentage of matched points required for two objects to be matched, respectively. $\mathds{1}$ is the indicator function that returns 1 if a distance $d_i$ is smaller than $np\_max\_distance$ and 0 otherwise. Thus, two objects are merged if the fraction of the number of pairwise distances smaller than $np\_max\_distance$ and the number of all pairwise distances is larger than $np\_min\_percentage$. The pipeline sets $np\_min\_percentage$ to 0.1 requiring two objects to have at least 10~\% of points within $np\_max\_distance$ meters. Thus, whether two objects are merged is mainly controlled by setting $np\_max\_distance$. Figure~\ref{fig:4} (\textbf{A}) shows three matched non-pipe objects from a total of 16 images using the approach outlined in this section.

\subsubsection{Pipe objects}\label{sec:pipe}

\begin{figure}[!ht]
    \begin{center}
        \includegraphics{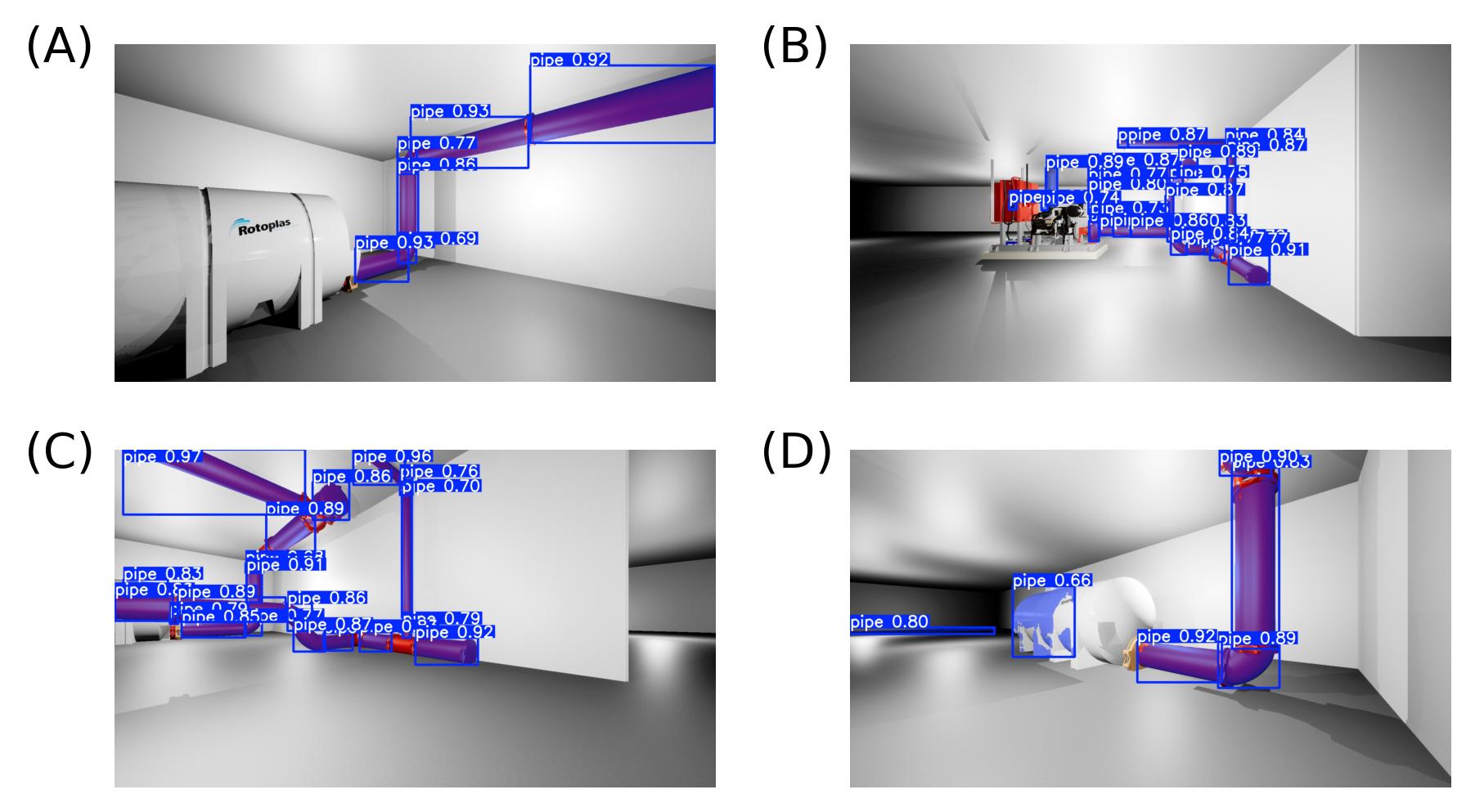}
    \end{center}
    \caption{Instance segmentation of pipe objects. Bounding boxes indicate detected objects with class label and class confidence score, while segmented objects are shaded blue. (\textbf{A}) -- (\textbf{D}) show different perspectives of hydraulic system 1.}\label{fig:5}
\end{figure}

The pipeline uses the YOLOv8 model \citep{redmon2016} to predict instance segmentations for individual pipe elements. This model is pre-trained on the instance segmentation images of the COCO data set \citep{lin2014}. Since the instance segmentation task is more complex compared to the pose estimation task, the largest architecture of YOLOv8 and a data set consisting of 1,030 images are applied. The pre-trained model is fine-tuned with 280 real-world RGB and 750 Unreal Engine images for 200 epochs and optimized using Adam with a learning rate of 10\textsuperscript{-4}. Figure~\ref{fig:5} shows predictions for four images. Similar to the model used in Section~\ref{sec:nonpipe}, each prediction consists of a bounding box with class label and class confidence score. However, instead of keypoints the model predicts masks for the pipe elements (shaded in blue). Predictions are not perfect (nor are they expected to be) and are characterised by some false negatives, where pipe elements are not detected, e.g., in Figure~\ref{fig:5} (\textbf{C}), and some false positives, where parts of the background or other objects are falsely predicted as a pipe, e.g., in Figure~\ref{fig:5} (\textbf{D}). There are also cases for which only parts of the pipe are detected or single pipe elements are split into two or more instances. The aim of the pipeline modules downstream from the instance segmentation are to fill in the missing pipe elements and to discard false detections in the generation of the graph.

\begin{figure}[!ht]
    \begin{center}
        \includegraphics{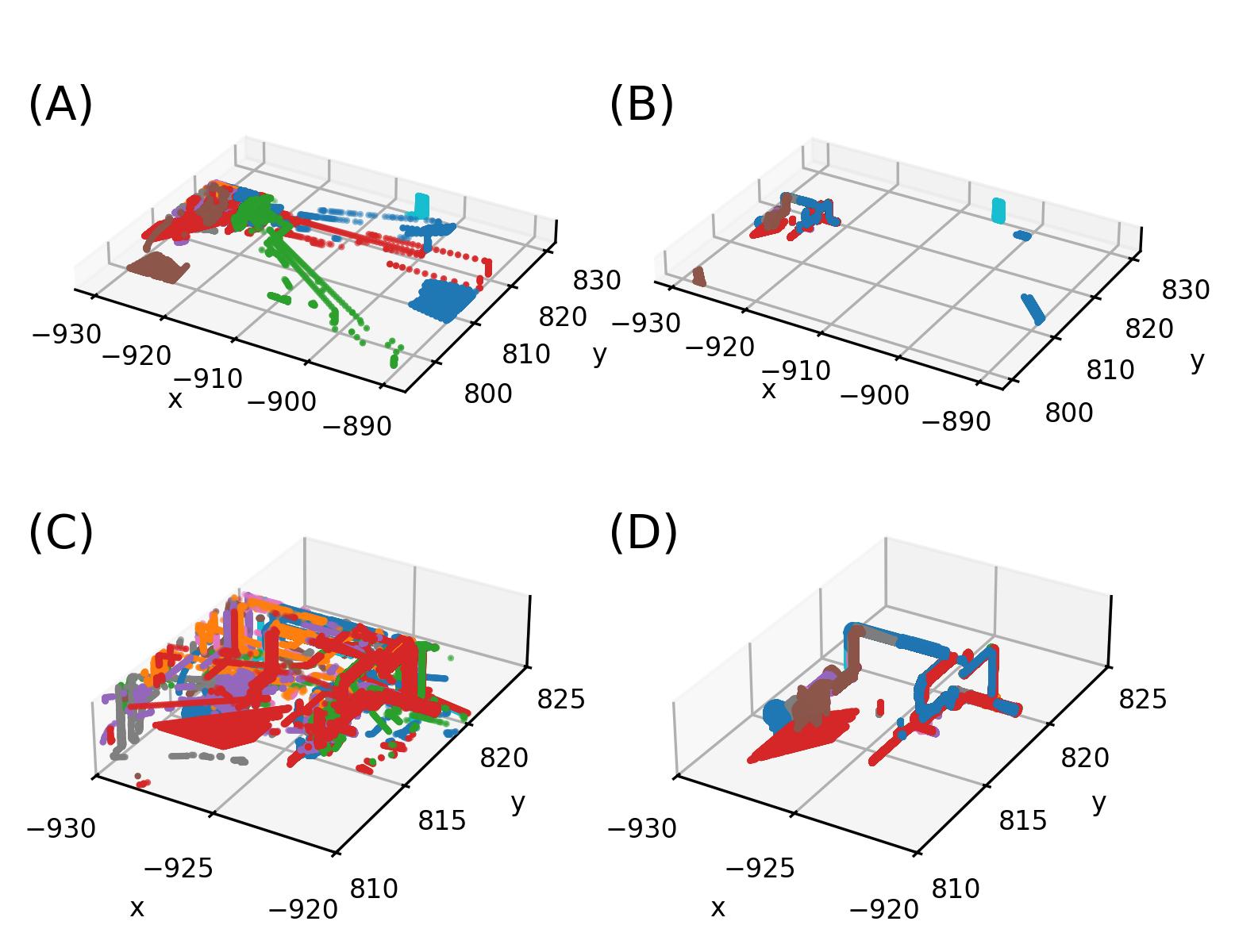}
    \end{center}
    \caption{Cleanup of pipe segmentations projected to world view. (\textbf{A}) shows the initial segmentations, while (\textbf{B}) shows segmentations after cleanup. (\textbf{C}) and (\textbf{D}) are a zoomed-in version of (\textbf{A}) and (\textbf{B}). Each color indicates segmentations from an individual image.}\label{fig:6}
\end{figure}

The detections of YOLOv8 are cleaned up by removing the outer two pixels of each mask as they oftentimes contain pixels from the background. The masks are then projected into the three-dimensional world view, where they are further cleaned with a combination of DBSCAN \citep{ester1996, schubert2017} and statistical outlier removal to discard any voxels that belong to the background or objects other than the specific pipe contained within the mask (similar to the process described in Section~\ref{sec:nonpipe}). In this step, the projection to three-dimensional space through the use of the depth data is instrumental to enable the differentiation between the pipe object and the background/other objects. The pinhole camera model is used for this projection \citep{hartley2003}. Finally, other non-pipe objects, such as the pump, also contain pipes that are detected by the model. However, in this case we do not require these pipes for our final relational graph as they are part of the non-pipe object. Thus, the final step of the 'Detection cleanup' module removes masks that are part of the previously matched non-pipe objects as the arrow from the 'Matching' module of the 'Non-pipe objects' branch to the 'Detection cleanup' module of the 'Pipe objects' branch in Figure~\ref{fig:2} indicates. Pipes that overlap with any of the non-pipe objects are eliminated from the data set. Figure~\ref{fig:6} shows the cleanup process from its starting point in (\textbf{A}) to its finished product in (\textbf{B}), where (\textbf{C}) and (\textbf{D}) show are second view zoomed in on the pipe network.

\begin{figure}[!ht]
    \begin{center}
        \includegraphics{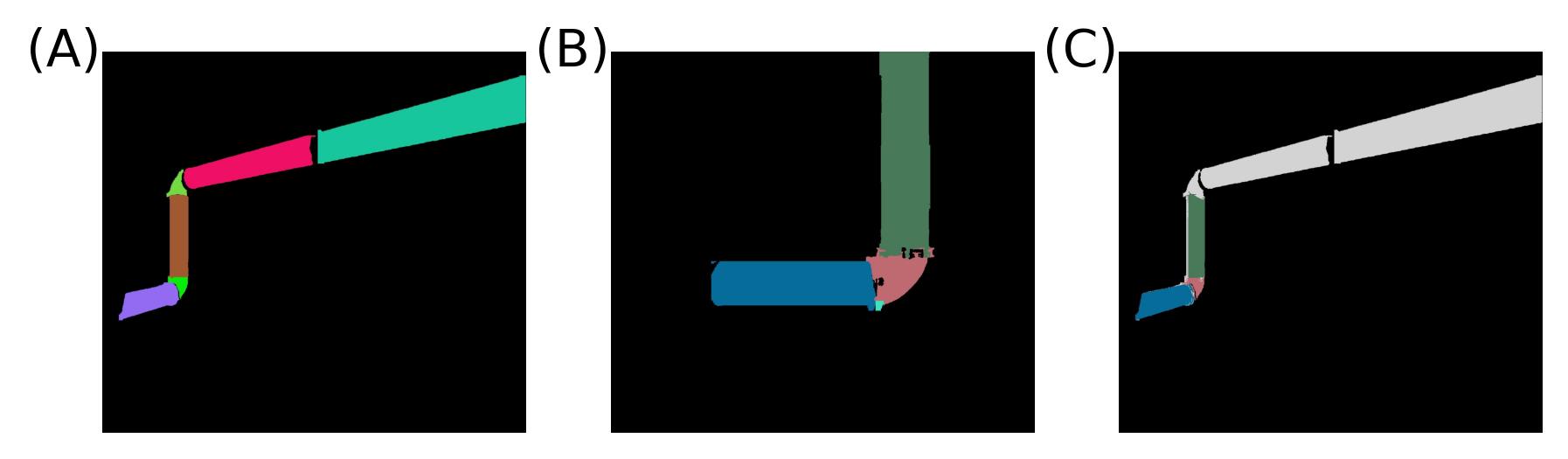}
    \end{center}
    \caption{Projection of segmented pipes. (\textbf{A}) is the target image on which segmented pipes of the source image (\textbf{B}) are projected. (\textbf{C}) shows the projected segmentations from (\textbf{B}) in color on the segmentations of (\textbf{A}) in gray.}\label{fig:7}
\end{figure}

After cleaning the detections, the remaining pipe masks are matched across the images. The matching process consists of two steps. First, each mask is projected onto all other images and the intersection with the masks of these images is computed. This is done for all masks and all images. This process is depicted for one image pair in Figure~\ref{fig:7}, where (\textbf{A}) shows the target image on which masks of the source image (\textbf{B}) are projected. (\textbf{C}) shows this projection with pipe elements of the target image now displayed in gray.  (\textbf{C}) shows that at least three pipe elements of the source image are projected onto pipe elements of the target image. This will result in a positive intersection area for these pipe element pairs indicating possible matches. Projections are based on the pinhole camera model \citep{hartley2003}. Second, to validate that the overlapping masks are actual matches, they are projected into three-dimensional world view, voxels are down-sampled to 1 cm\textsuperscript{3} and DBSCAN \citep{ester1996, schubert2017} is used to find distinct clusters that are at least 10 cm apart. Each of these clusters is subsequently considered an individual object. Figure~\ref{fig:4} (\textbf{B}) shows the resulting matched pipe objects from all 16 images of hydraulic system 1.

\begin{figure}[!ht]
    \begin{center}
        \includegraphics{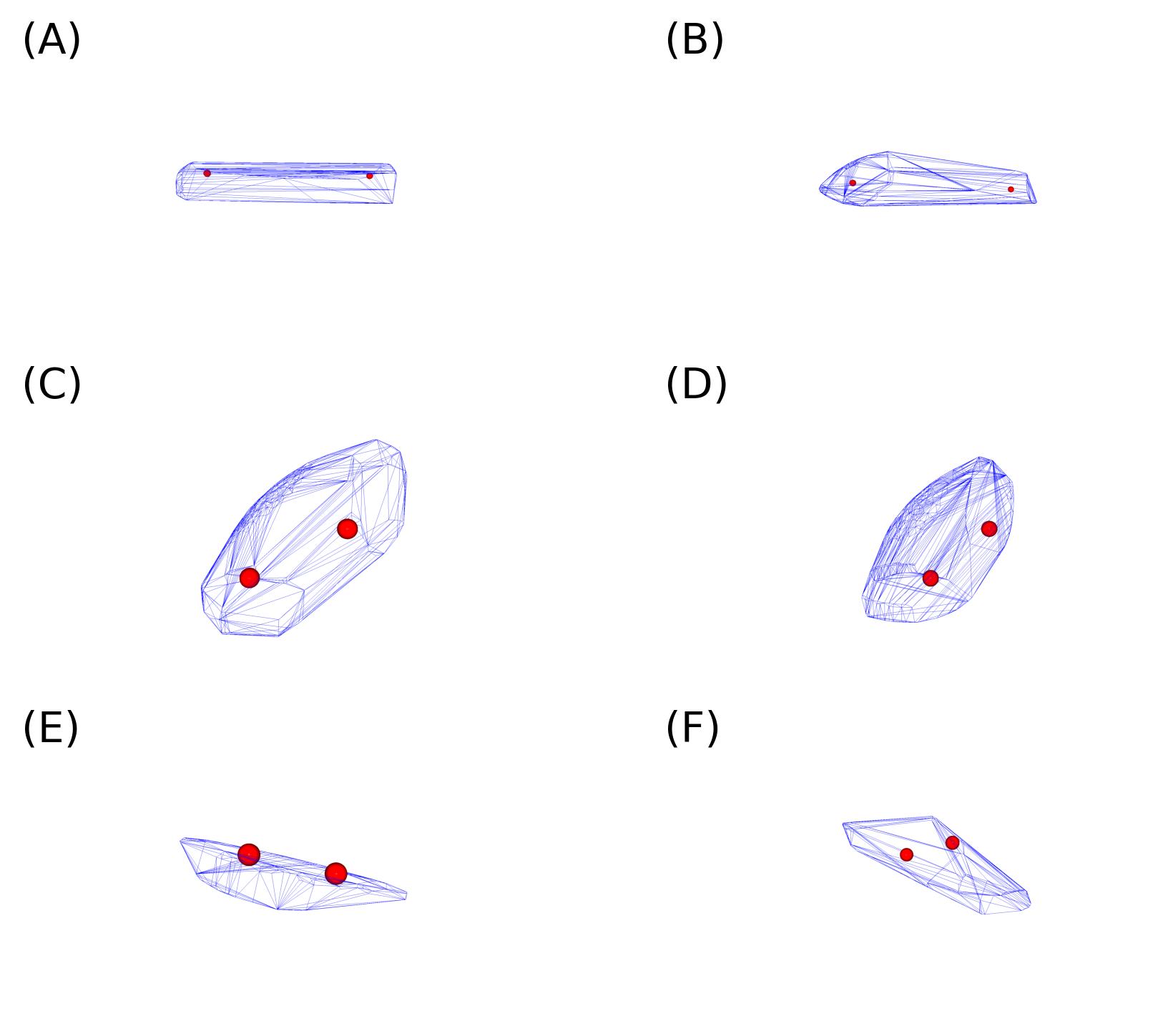}
    \end{center}
    \caption{Endpoint approximation of pipe elements. Blue lines are a wireframe representation of the pipe elements while red spheres indicate approximated endpoints. (\textbf{A}) and (\textbf{B}) show approximations for straight pipe elements, (\textbf{C}) and (\textbf{D}) for bent pipe elements, and (\textbf{E}) and (\textbf{F}) for ambiguous pipe elements.}\label{fig:8}
\end{figure}

While the endpoints of the non-pipe objects are predicted by YOLO as outlined in Section~\ref{sec:nonpipe}, the YOLOv8 model used for instance segmentation does not predict keypoints. Thus, an approximation of the endpoints for the pipe objects is required. Pipe objects can be classed into two categories: straight pipes and non-straight pipes, such as bent pipes and T-fittings. We differentiate between these classes by computing a rotated bounding box around the objects and comparing its longest side against its shorter sides. If the ratios of $\frac{\text{length of long side}}{\text{length of short side}}$ are larger than hyperparameter $p\_threshold \in [0, \infty)$, the pipe is classed as straight. If the ratios are equal or smaller than this hyperparameter, the pipe is classed as non-straight. Voxels are binned along the longest side of the bounding box and the mean of all voxels within the first and the last bin are used as the two endpoints of the pipe. Figure~\ref{fig:8} (\textbf{A}) and (\textbf{B}) show wireframes of the hull of two straight pipes in blue and the approximated endpoints as red spheres. For the non-straight pipes, it is generally more challenging to find an appropriate heuristic that gives satisfying results. Thus, we begin by computing the centroid of the object and place it towards the most relevant neighboring object. In detail, we find the object that has the highest number of voxels within $p\_max\_distance \in (0, \infty)$ meters of the hull of the pipe object and use the linear combination $(1-p\_w)\times centroid + p\_w \times neighbor$ to find the final position of the endpoint. $p\_max\_distance$ is the maximal distance in meters around the pipe objects hull in which other objects are considered and $p\_w \in [0, 1]$ controls how far a centroid is pulled towards the other objects. The pipeline sets $p\_w = 0.3$ which slightly nudges the endpoints towards the objects neighboring in close proximity. This process is executed twice for each centroid yielding two endpoints for each pipe object. Figure~\ref{fig:8} (\textbf{C}) and (\textbf{D}) show two examples of two 90° bents. While this method provides satisfying results for most objects, it struggles with pipe objects that have an ambiguous shape (Figure~\ref{fig:8} (\textbf{E}) and (\textbf{F})) as discussed in detail as part of the limitations in Section~\ref{sec:discussion}.

\subsubsection{Connection prediction and Graph generation}\label{sec:graph}

\begin{figure}[!ht]
    \begin{center}
        \includegraphics{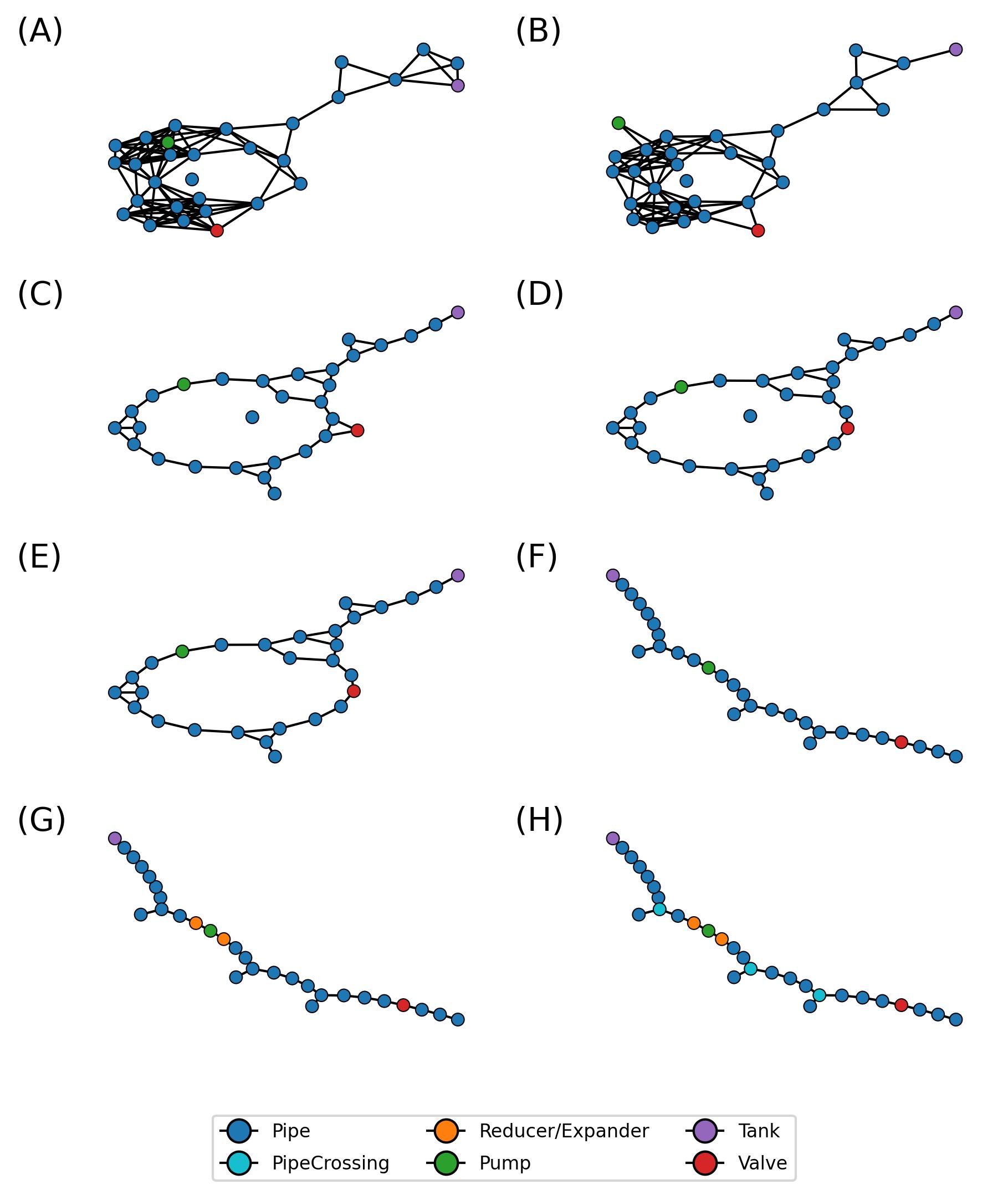}
    \end{center}
    \caption{Graph generation. (\textbf{A}) shows the initial graph based on distance between endpoints, and (\textbf{B}) -- (\textbf{H}) show graphs enforced by the individual rules discussed in Section~\ref{sec:graph}.}\label{fig:9}
\end{figure}

The processes for non-pipes (Section~\ref{sec:nonpipe}) and for pipes (Section~\ref{sec:pipe}) yield endpoints for each object that can be used to predict how individual objects are connected and how all objects that form an entire hydraulic system relate to each other. The whole scene can effectively be described as a relational graph in which objects are nodes and object connections are indicated by edges between these nodes. The idea of the graph generation process is to first create a initial graph with probable connections and then refine this graph by defining and enforcing a set of rules. This process is depicted in Figure~\ref{fig:9}.

The initial graph is generated by computing the pairwise distances between the endpoints of all objects and iteratively connecting objects, i.e., drawing edges between nodes weighted by their respective pairwise distance, starting with the smallest distance. Nodes are iteratively connected until all remaining distances between endpoints are larger than the hyperparameter $graph\_max\_distance \in (0, \infty)$. Two individual objects are restricted to a single connection. An example of an initial graph is shown in Figure~\ref{fig:9} (\textbf{A}).

The initial graph is then refined by defining a set of rules and enforcing them. This means that edges breaking at least one of the rules are deleted from the graph, beginning with the edge with the largest distance, until no further rule violation exists. For the hydraulic systems considered in this articles, the following rules are enforced:

\begin{itemize}
    \item[-] \textbf{Rule 1}: Pumps and valves are limited to a maximum of two connections, tanks to a maximum of one connection.
    \item[-] \textbf{Rule 2}: Pipes are limited to a maximum of three connections.
    \item[-] \textbf{Rule 3}: Neighbors of the same non-pipe object cannot be connected.
    \item[-] \textbf{Rule 4}: Single nodes are not allowed.
    \item[-] \textbf{Rule 5}: Cycles are not allowed.
    \item[-] \textbf{Rule 6}: Pipes directly connected to pumps are classed as Reducer/Expander.
    \item[-] \textbf{Rule 7}: Pipes with three neighbors are classed as PipeCrossing.
\end{itemize}

The graphs in Figure~\ref{fig:9} (\textbf{B}) through (\textbf{H}) display the results after enforcing each of the rules. While the initial graph (\textbf{A}) is convoluted and no clear system can be identified, the graph takes a more realistic shape with each step until its final shape is reached at (\textbf{F}). After that updates only address the type of the existing nodes. Straight and bend pipes are assigned type 'Pipe', T-fittings are assigned type 'PipeCrossing', and pipes adjacent to pumps are assigned type 'Reducer/Expander'. The diameter of the latter decreases or increase in the direction of the flow to facilitate proper functioning of the connected pump. The final graph is thoroughly discussed in Section~\ref{sec:system1}.

While these rules work well for the hydraulic system considered in this paper, they are not appropriate or sufficient for every type of structural critical infrastructure. However, the pipeline makes it straightforward to add new rules. Each rule is essentially a function that takes the current graph as the input and returns the adjusted graph as the output. The way in which the function alters the graph is defined by the rule itself. For Rule 1, for example, the function gathers all pumps and valves, counts the connections of each and deletes edges beginning with the largest distance for instances where one pump or valve has more than two connections. Adding a new rule requires defining a new function and inserting it at the appropriate position relative to the other rules. An advantage of implementing rules as simple functions is that their behaviour can easily be validated and scrutinized by feeding them a graph that breaks the specific rule and observing if the returned graph no longer includes rule violations. This makes the rules transparent and their behavior predictable. Thus, this approach is easy to interpret, particularly, when compared to methods relying on deep learning black boxes as introduced in Section~\ref{sec:relatedworks}.


\section{Results}\label{sec:results}

This section presents the results for two hydraulic systems generated by the Unreal Engine 5 \citep{epicgames2022a}, as discussed in Section~\ref{sec:images}. We compare the predicted graph against the ground truth in Figures~\ref{fig:10} and \ref{fig:11} and highlight differences through three-dimensional plots depicting the individual objects of the systems in Figure~\ref{fig:12}. Neighboring connected pipe objects are contracted in these figures to make the comparison, both qualitative and through computing the graph edit distance \cite{gao2010} as a quantitative metric, more convenient. Furthermore, we maintain the position of the pipes to enable downstream simulation. Hence, only the visual appearance of the graphs changes and no information about the actual 3D system is lost. This is shown in Figure~\ref{fig:12} where each individual object is retrieved from the graphs. Values of the hyperparameters used to generate these results are shown in Table~\ref{tab:1}. Section~\ref{sec:pipeline} describes all hyperparameters in detail and Section~\ref{sec:sensitivity} discusses how hyperparameters were selected and conducts a sensitivity analysis. Potential implications of these differences are addressed in Section~\ref{sec:discussion}.

\begin{table}
    \renewcommand{\arraystretch}{1.25}
    \centering
    \caption{Parameters used in the prediction of hydraulic systems 1 and 2.}\label{tab:1}
    \begin{tabular}{ l l | c c }
        Parameter & Module & System 1 & System 2 \\
        \hline\hline
        $np\_max\_distance$ & Non-pipe matching & 0.75 & 0.75 \\
        $p\_matching\_min\_overlap$ & Pipe matching & 0.70 & 0.70 \\
        $p\_threshold$ & Pipe endpoint estimation & 2.00 & 2.00 \\
        $p\_max\_distance$ & Graph generation & 1.50 & 1.50 \\ 
        $graph\_max\_distance$ & Graph generation & 1.50 & 1.50
    \end{tabular}
\end{table}

\subsection{Hydraulic system 1}\label{sec:system1}

\begin{figure}[!ht]
    \begin{center}
        \includegraphics{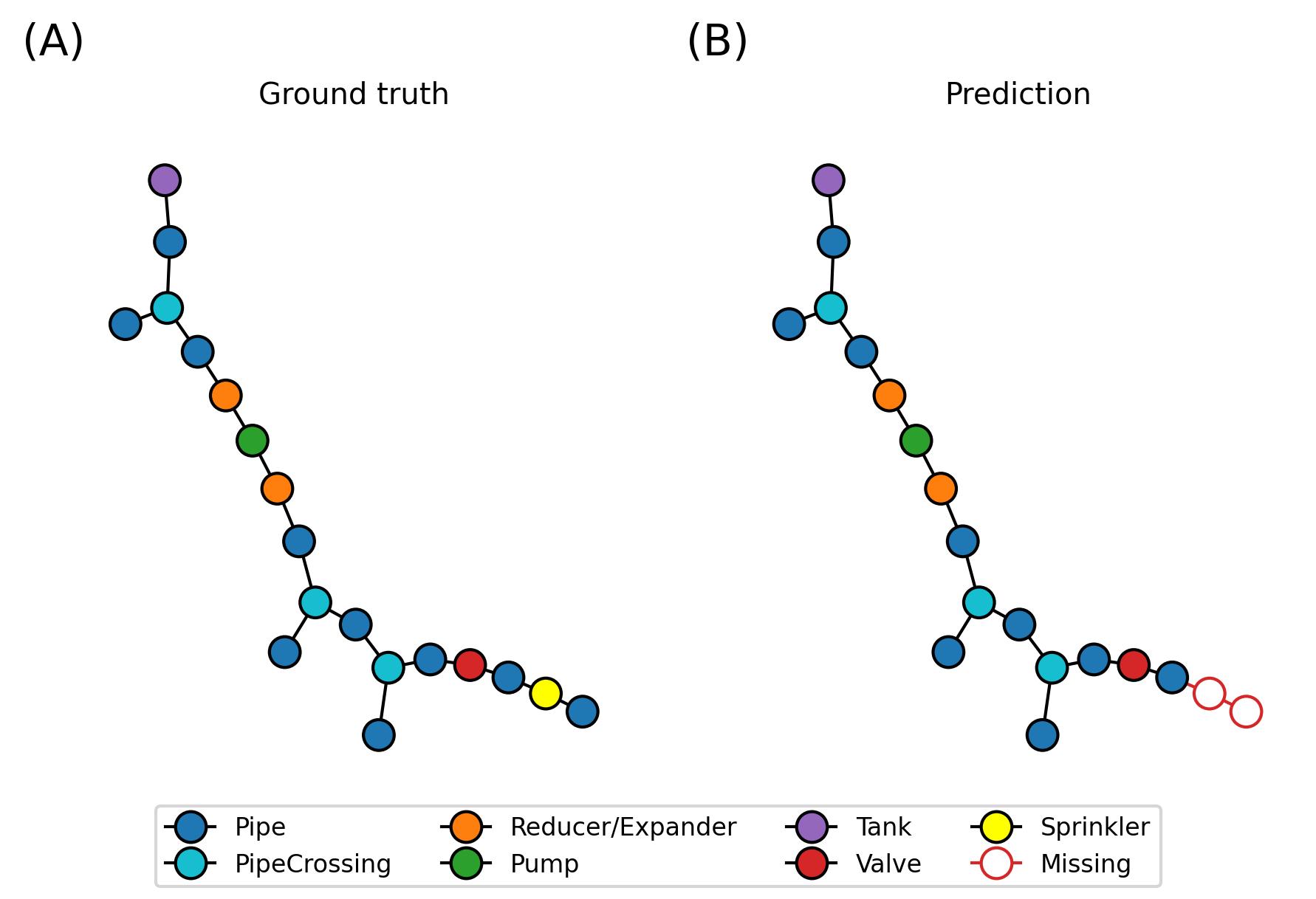}
    \end{center}
    \caption{Comparison of ground truth (\textbf{A}) to predicted graph (\textbf{B}) for hydraulic system 1.}\label{fig:10}
\end{figure}

Hydraulic system 1 consists of one tank, one pump, one valve and one sprinkler as well as three T-fittings as displayed in the ground truth graph in Figure~\ref{fig:10} (\textbf{A}). Figure~\ref{fig:10} (\textbf{B}) shows the graph predicted by the pipeline detailed in Section~\ref{sec:pipeline} with errors indicated by red edges. The overall shape of the graphs is almost identical. The three T-fittings (displayed as 'PipeCrossing') are present in both graphs and the graph starts with a tank on one end, followed by the pump and valve and pipe elements in between each of the non-pipe objects. Solely the sprinkler is not included. However, this is not a shortcoming of the pipeline itself but rather the sprinklers in the images of the specific data set are too small to be detected by the fine-tuned YOLOv8 model \citep{redmon2016}. The graph edit distance \cite{gao2010} using a cost of 1 for deletion and insertion and a cost of 2 for substitution equates to 4. This indicates a high similarity between the ground truth and the prediction. Figure~\ref{fig:12} (\textbf{A}) shows some errors in the prediction of the individual pipe objects. Particularly, there are instances where two subsequent pipe objects are falsely predicted as one pipe object: The gold pipe parallel to the y-axis at ground level at (-924, 819) and the olive pipe parallel to the y-axis at ceiling level at (-922, 820), respectively. Both pipes combine a straight pipe and a 90° bent in one pipe object. Moreover, the plot shows that the tank (bright green object at (-928, 817)) is approximated in the correct location but is not detected entirely missing a significant part towards the x-axis. Lastly, a pipe object at ceiling level at (-924, 819) is partly missing. Implications of these error and mitigation are discussed in Section~\ref{sec:discussion}.

\subsection{Hydraulic system 2}\label{sec:system2}

\begin{figure}[!ht]
    \begin{center}
        \includegraphics{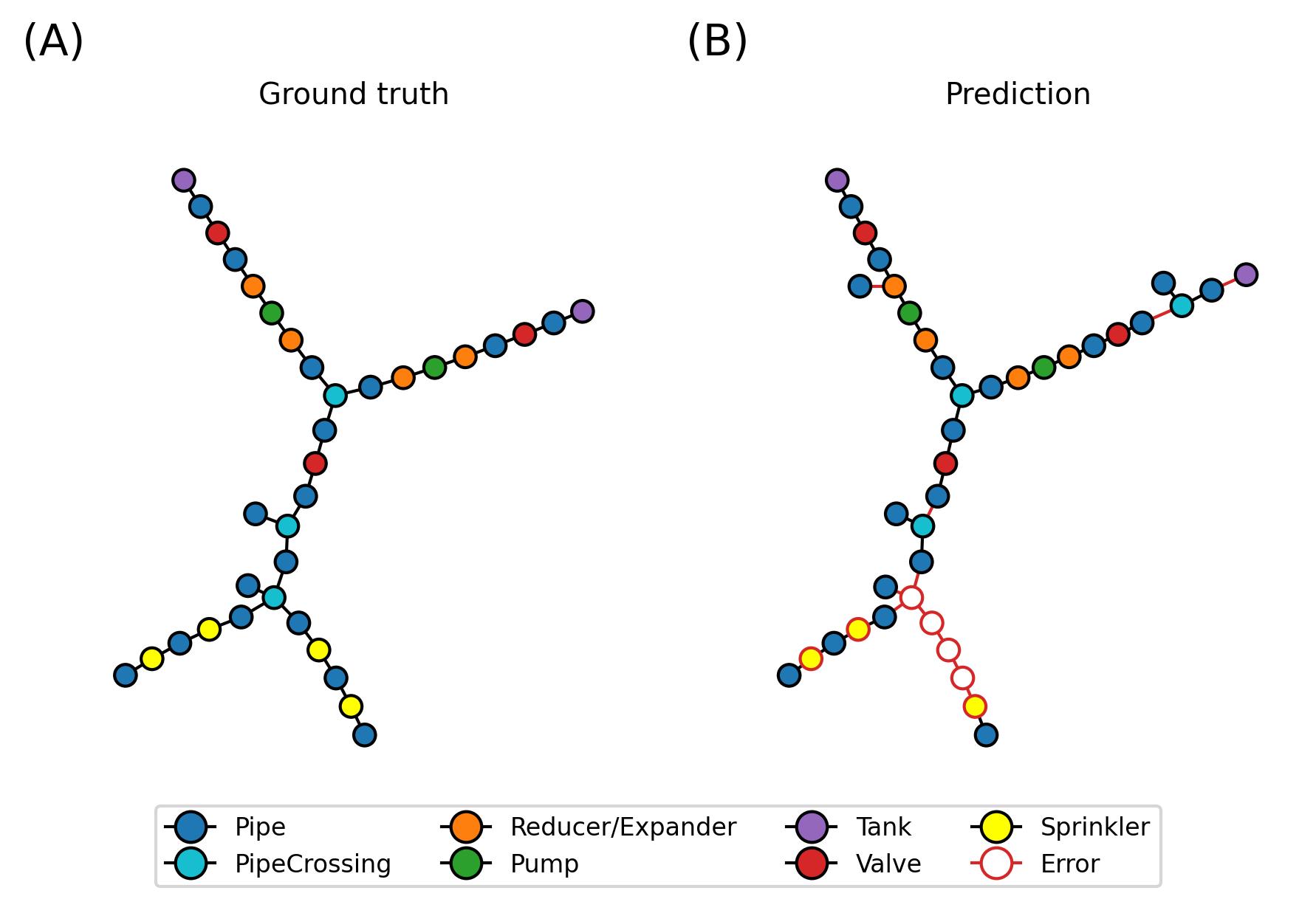}
    \end{center}
    \caption{Comparison of ground truth (\textbf{A}) to predicted graph (\textbf{B}) for hydraulic system 2.}\label{fig:11}
\end{figure}

\begin{figure}[!ht]
    \begin{center}
        \includegraphics{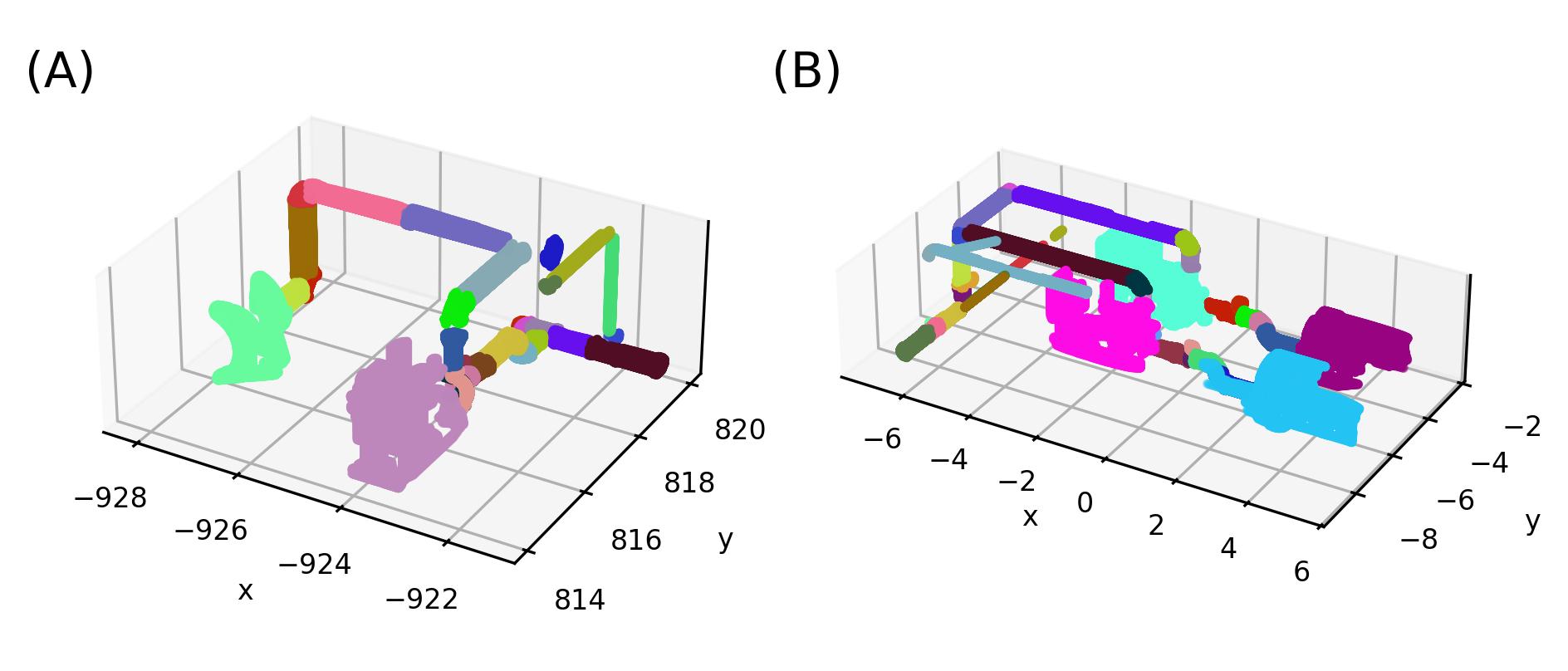}
    \end{center}
    \caption{Prediction of individual objects for hydraulic system 1 (\textbf{A}) and hydraulic system 2 (\textbf{B}).}\label{fig:12}
\end{figure}

Hydraulic system 2 is more complex than system 1. It consists of two tanks, two pumps, three valves and four sprinklers separated by various pipes and T-fittings. The system has an X-shape with the tanks and pumps on one side and the sprinklers on the other as shown in the ground truth in Figure~\ref{fig:11} (\textbf{A}). The prediction (\textbf{B}) of the side with the pumps and tanks only shows a minor difference in the number of predicted pipe elements between the valves and the pumps. This difference is due to the detection of extra pipe elements by YOLOv8 \citep{redmon2016}, as discussed in Section~\ref{sec:pipe}, and the consequent incorrect matching within the pipeline. In Figure~\ref{fig:12}, these errors manifest, for example, as the small additional pink pipe element at (0, -6). The opposite side with the sprinkler system of the graph in Figure~\ref{fig:11} (\textbf{B}) is not connected to the graph due to an undetected thin vertical pipe at (-6, -8) in Figure~\ref{fig:12} (\textbf{B}). See Figure~\ref{fig:1} (\textbf{B}) for comparison, where the thin vertical pipe can be seen on the left. Similar to system 1, the sprinklers objects were too small to be detected by YOLOv8 and thus are neither included in the graph nor the three-dimensional representation of the system. The GED \cite{gao2010} for the entire hydraulic system 2 is 22. Considering that the reason for the missing sprinklers is not the pipeline itself but rather the detection model which is just an input to the pipeline and can be replaced with any prediction model, disregarding the errors caused by the missing sprinklers might be a more accurate metric. In this case the GED drops to 10. 

\section{Sensitivity analysis and runtimes}\label{sec:sensitivity}

In this section, we want to outline how good values for the hyperparameters listed in Table~\ref{tab:1} can be selected before analyzing the sensitivity of the outcome to changes in these hyperparameters. 

\begin{figure}[!ht]
    \begin{center}
        \includegraphics{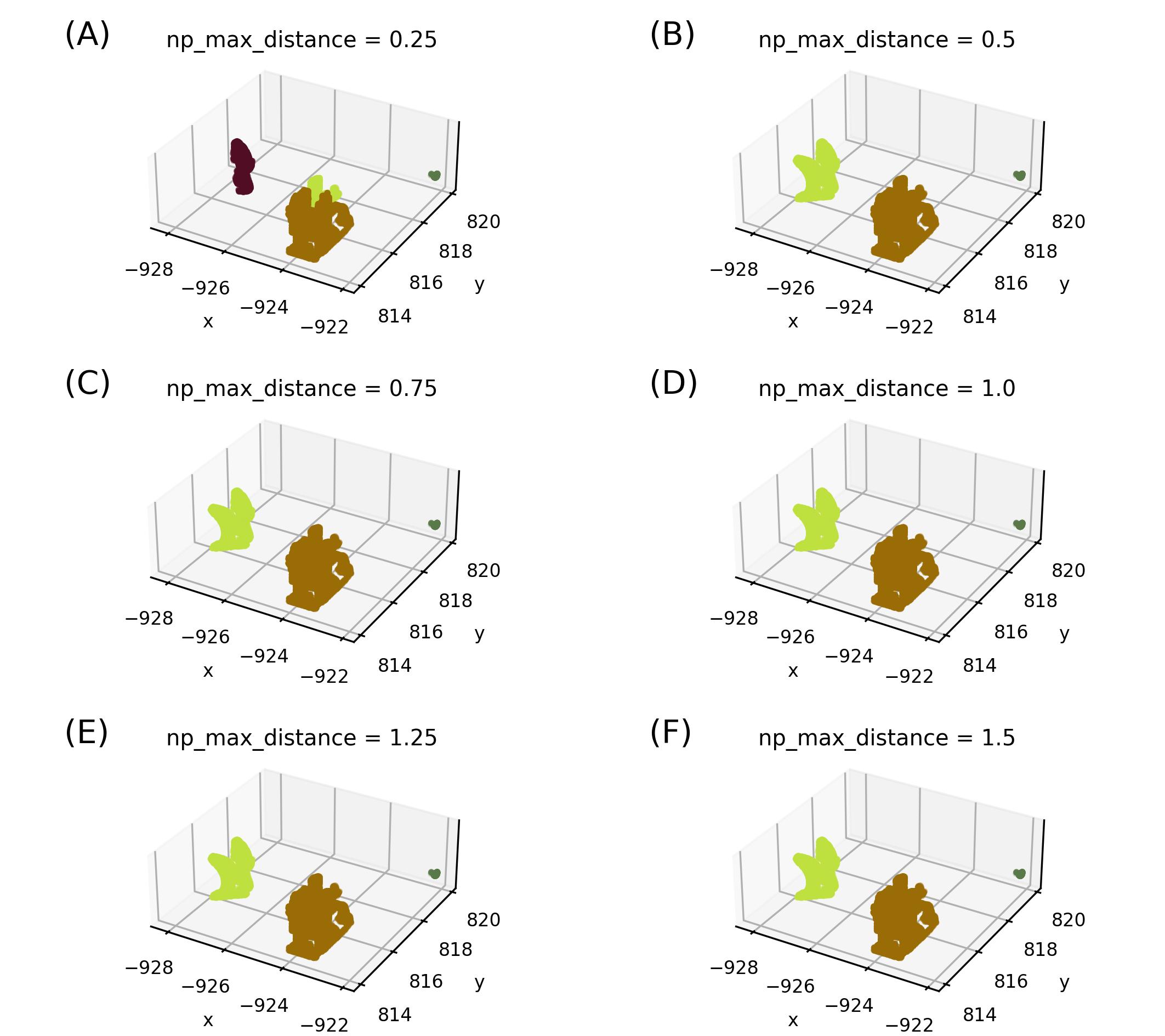}
    \end{center}
    \caption{Tuning of hyperparameter $np\_max\_distance$. Orange points indicate hyperparameters used for hydraulic system 1. (\textbf{A}) $np\_max\_distance = 0.25$. (\textbf{B}) $np\_max\_distance = 0.5$. (\textbf{C}) $np\_max\_distance = 0.75$. (\textbf{D}) $np\_max\_distance = 1.0$. (\textbf{E}) $np\_max\_distance = 1.25$. (\textbf{F}) $np\_max\_distance = 1.5$.}\label{fig:14}
\end{figure}

\begin{figure}[!ht]
    \begin{center}
        \includegraphics{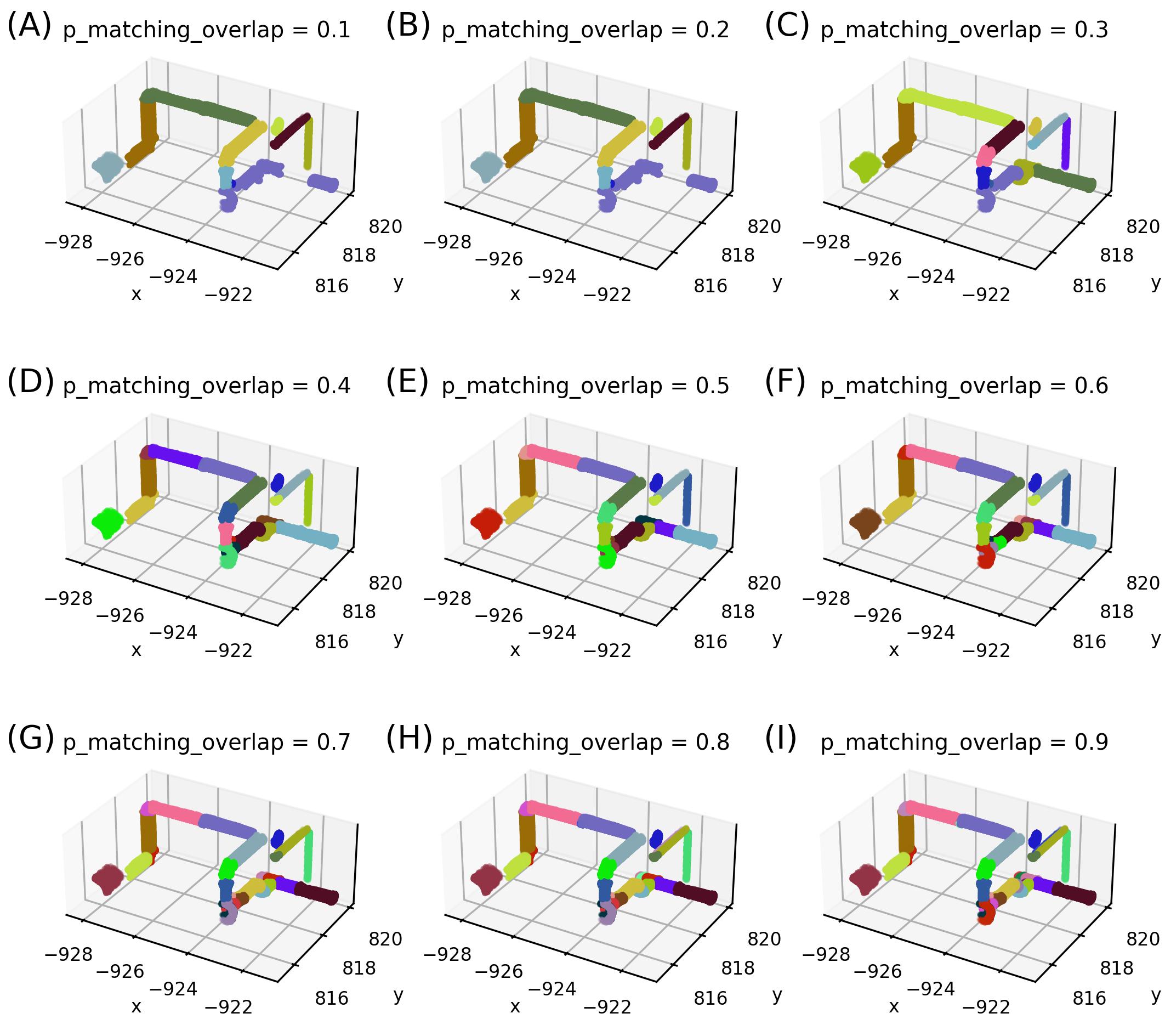}
    \end{center}
    \caption{Tuning of hyperparameter $p\_matching\_overlap$. Orange points indicate hyperparameters used for hydraulic system 1. (\textbf{A}) $p\_matching\_overlap = 0.1$. (\textbf{B}) $p\_matching\_overlap = 0.2$. (\textbf{C}) $p\_matching\_overlap = 0.3$. (\textbf{D}) $p\_matching\_overlap = 0.4$. (\textbf{E}) $p\_matching\_overlap = 0.5$. (\textbf{F}) $p\_matching\_overlap = 0.6$. (\textbf{G}) $p\_matching\_overlap = 0.7$. (\textbf{H}) $p\_matching\_overlap = 0.8$. (\textbf{I}) $p\_matching\_overlap = 0.9$.}\label{fig:15}
\end{figure}

An advantage of the modular approach presented in Figure~\ref{fig:2} is that we can tune the hyperparameters of each module individually. For the none-pipe objects, the only hyperparameter to set is $np\_max\_distance$ which controls the merging of objects detected in multiple images. Particularly, it provides the maximal distance in meters that detections can be apart in three-dimensional space to be combined into one object. Intuitively, this parameter can be set to just below the minimal pairwise distance between all non-pipe objects. Figure~\ref{fig:14} shows the final none-pipe objects for different values of $np\_max\_distance$. Evidently, if the parameter is too small, one object might be falsely split into multiple objects as is the case for $np\_max\_distance = 0.25$ in Figure~\ref{fig:14} (\textbf{A}). If the parameter is set too large, multiple distinct object might be combined into a single object, however, this is not the case up to the maximum considered distance of 1.5 m in Figure~\ref{fig:14} (\textbf{F}). Thus, setting the value to a distance that is just below the minimal pairwise distance between all non-pipe objects is advisable.

The only hyperparameter that requires tuning for processing and matching the pipe objects is $p\_matching\_overlap$ which controls the minimal required overlap of two masks to be matched together as shown in Figure~\ref{fig:7}. Figure~\ref{fig:15} shows results for a range of $p\_matching\_overlap$ values, where each color indicates one pipe object. If the parameter is set too low, many pipe objects will erroneously be matched together. If the parameter is set too high, individual pipe elements might be split into multiple objects. When setting this parameter, it is easiest to plot a range of different values and select the highest value for which individual pipe elements are not split into multiple objects. In this case $p\_matching\_overlap = 0.7$ meets this criteria. This process is feasible as the pipe matching process takes an average of 14.55 s to complete (as discussed later in this section) and thus enables rapid experimentation.

When setting the remaining three hyperparameters $p\_threshold$, $p\_max\_distance$ and $graph\_max\_distance$ the same process as before can be applied as approximating the endpoints and generating the graph only takes 1.64 s. We chose the graph in Figure~\ref{fig:10} as the best performing solution which uses $p\_threshold = 2.0$, $p\_max\_distance = 1.5$ and $graph\_max\_distance = 1.5$. To make this approach of setting the hyperparameters feasible for large scenes, hyperparameter tuning can either be performed on a small area of the scene and then used for the full scene, or it can be performed on one scene and then transferred to another. We opted for the latter option and tuned the hyperparameters on system 1 and applied the pipeline using the same hyperparameters to system 2.

\begin{figure}[!ht]
    \begin{center}
        \includegraphics{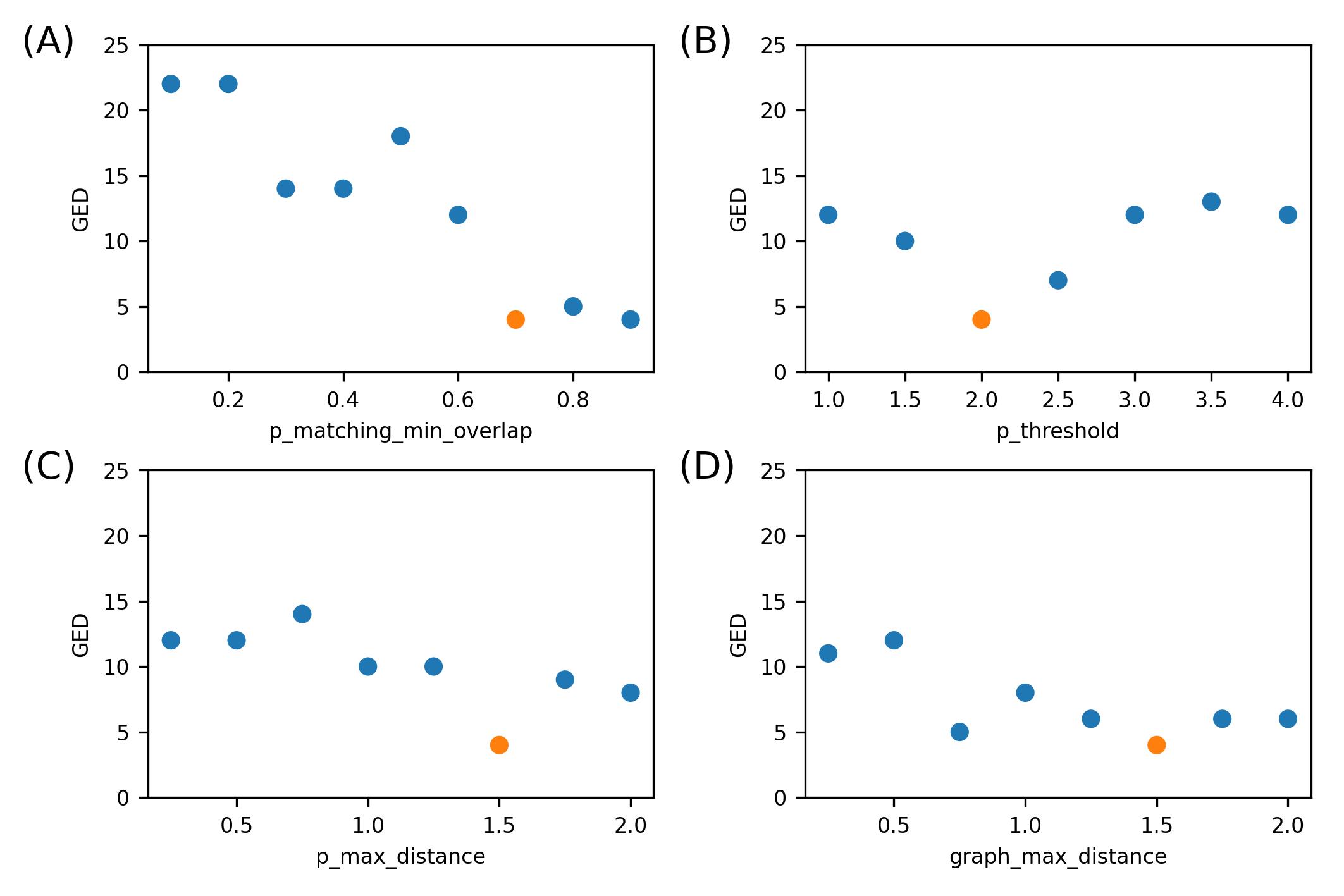}
    \end{center}
    \caption{Sensitivity analysis of pipeline hyperparameters. Orange points indicate hyperparameters used for hydraulic system 1. (\textbf{A}) -- (\textbf{D}) show the GEDs resulting from different values of $p\_matching\_min\_overlap$, $p\_threshold$, $p\_max\_distance$ and $graph\_max\_distance$, respectively.}\label{fig:13}
\end{figure}

Furthermore, we conducted a sensitivity analysis of the hyperparameter to give further guidance on how changes affect the results. Figure~\ref{fig:13} provides the graph edit distance (GED)  \citep{gao2010} for four hyperparameters over a range of different values, where the orange circles display the hyperparameters used for the results presented in Section~\ref{sec:results}. The GED is computed with a insertion and deletion cost of 1 and a node substitution cost of 2. We only adjust the value for the displayed hyperparameter while the others remain as given in Table~\ref{tab:1}. The analysis shows that values above $p\_matching\_overlap = 0.6$ and $graph\_max\_distance = 1.0$ result in similar GEDs, while $p\_threshold$ and $p\_max\_distance$ are more sensitive to changes displaying optimal values at 2.0 and 1.5, respectively. This indicates that it is advisable to set $p\_matching\_overlap$ and $graph\_max\_distance$ towards the upper end of the parameter range while results profit from a more precise tuning of parameters $p\_threshold$ and $p\_max\_distance$ following the process detailed above.

Finally, we present the runtimes of the individual modules of Figure~\ref{fig:2} in Table~\ref{tab:2}. These runtimes correspond to the 29 runs of the sensitivity analysis. The mean time it takes to run the entire pipeline once is 785.68 s with a standard deviation of 3.85 s. However, results of earlier modules can be saved and reused for subsequent module such that for each module previous modules do not have to be run again. For the hyperparameter tuning of the parameters in the modules Pipes endpoints and Graph generation this means that previous modules do not have to rerun again and various hyperparameter combinations for $p\_threshold$, $p\_max\_distance$ and $graph\_max\_distance$ can be tested as they only take 1.63 s on average to evaluate. This enables rapid experimentation and makes the process of setting the hyperparameters as described above feasible and practical. Experiments were run in parallel on an Intel Xeon Platinum 8260 CPU and 192 GB RAM.

\begin{table}
    \renewcommand{\arraystretch}{1.25}
    \centering
    \caption{Runtime statistics in seconds from applying the pipeline 29 times with different hyperparameters to hydraulic system 1.}\label{tab:2}
    \begin{tabular}{ l | c r r r r r }
        Process & n & Median & Mean & Std. deviation & Minimum & Maximum \\
        \hline\hline
        Preprocessing & 29 & 67.02 & 67.10 & 0.52 & 66.16 & 68.61 \\
        None-pipes object detection & 29 & 3.06 & 3.09 & 0.11 & 3.01 & 3.60 \\
        None-pipes location approximation & 29 & 90.34 & 90.25 & 1.55 & 86.93 & 92.51 \\
        None-pipes matching & 29 & 15.49 & 15.52 & 0.14 & 15.33 & 15.90 \\
        Pipes object detection & 29 & 16.17 & 16.18 & 0.13 & 15.93 & 16.54 \\
        Pipes cleanup & 29 & 14.56 & 14.54 & 0.20 & 14.04 & 14.84 \\
        Pipes matching preprocess & 29 & 553.24 & 552.02 & 3.58 & 544.70 & 557.84 \\
        Pipes matching & 29 & 14.36 & 14.55 & 0.79 & 13.68 & 16.74 \\
        Pipes endpoints & 29 & 1.14 & 1.05 & 0.27 & 0.42 & 1.48 \\
        Graph generation & 29 & 0.58 & 0.58 & 0.09 & 0.41 & 0.99 \\
        \hline
        Total & 29 & 775.47 & 774.86 & 3.85 & 767.86 & 782.43
    \end{tabular}
\end{table}


\section{Discussion}\label{sec:discussion}

The results presented in Section~\ref{sec:results} show that a combination of data acquisition via photogrammetry, object detection on images and graph generation with an user-defined set of rules is able to generate graphs that are close to the ground truth for two hydraulic systems. While the detection and matching of relevant objects and the prediction of their relations and connections works well, the following points should be noted. Firstly, pipe elbows and T-fittings are treated identical and for both exactly two endpoints are computed, although T-fittings have three connection points in reality. Rule 2, described as part of the graph generation process in Section~\ref{sec:graph}, limits the number of connections of each pipe to a maximum of three. This enables the method to identify the T-fittings although they are not detected as such initially. Secondly, the location approximation of non-pipe objects as described in Section~\ref{sec:nonpipe} uses some crude heuristics to prevent the need of time-consuming labelling required for instance segmentation of the objects. Albeit that the heuristics are crude, they are able to approximate the location of the non-pipe objects sufficiently as shown in Figure~\ref{fig:12}. The pumps are the most complex objects and are clearly identifiable in both images. Only the tank in Figure~\ref{fig:12} (\textbf{A}) is hard to identify. However, this is because the side of the tank towards the x-axis is not included in any of the images and thus cannot be detected. Thirdly, we use a simple approximation for the pipe endpoints as presented in Section~\ref{sec:pipe}. The results show however that this approximation is sufficient for connecting most pipe elements. Only the approximation of many small pipe elements in close proximity as, for example, around the valve between the pumps and tanks in system 2 is challenging. For the use in digital twins and simulations, it is likely that the individual pipe elements between two non-pipe objects will be contracted into a single pipe object as it is the case for the graphs in Figure~\ref{fig:11} and \ref{fig:12}. Hence, the pipe objects between two non-pipe objects are more relevant than how they are connected. While this means that the pipe predicted by our method will be correctly represented in the simulation model, it would be preferable to have a more accurate way of approximating endpoints for smaller pipes that will make the prediction of connections more robust. Section~\ref{sec:pipe} showed that the approximation performs poorly on pipe elements that are of ambiguous shape, i.e., they are not identifiable as straight or bent pipe elements. A more precise approximation could alleviate this issue. Fourthly, Section~\ref{sec:results} showed that some subsequent pipe elements were mistakenly matched together resulting in fewer pipe objects than actually exist. Although technically incorrect, this makes no difference for the functioning of the digital twin if we apply the same logic as for the previous point assuming that the matched pipe elements are actually connected in reality. Consider the case where two subsequent pipe elements are detected as two distinct objects, and the case where the same two pipe elements are falsely matched into a single object by the pipeline. When the 3D locations of the pipes that are stored in the graph are transferred into three-dimensional representation, the model will be the same. The only difference is that for the first case the pipes are connected when transferring the objects into the 3D representation while for the second case the pipes are already connected before the transfer. The model on which the downstream simulations are based will be identical and so should the simulation accuracy. Lastly, we want to point out that the performance of the proposed pipeline of this article is heavily influenced by the quality of the images and the models used for object detection and instance segmentation. Low-resolution images might not be accurate enough to accurately predict all relevant objects and segmentation masks might be ambiguous resulting in the inclusion of other objects or portions of the background. It can also make training more challenging for the model and prevent it from learning the most important features. Errors made by the prediction model, e.g., due to shortcomings in the model itself or poor image quality, are challenging to be reversed downstream in the pipeline. While the multiple perspectives of the images provide multiple opportunities to detect each object, some objects, such as the sprinklers or the thin vertical pipe in system 2, were not detected by the fine-tuned YOLOv8 model \citep{redmon2016} and thus were not included in the final graphs. To mitigate this issue, (a) the detection model could be improved with more training data focusing on areas with poor performance, (b) a different model could be explored, or (c) more images of the hydraulic system could be included that show the challenging objects more clearly.

Considering the limitations of the approach presented in this article, future work should address the approximation of the location of the non-pipe objects and of the pipe endpoints. One possible solution could be the skeletonization of the pipe elements as discussed in \cite{Alex2025} and \cite{meyer2023}. Furthermore, although the object detection achieved good results overall, there were issues in detecting smaller objects, such as sprinklers and thin pipes. It could be investigated if this can be improved by collecting more images where small objects are depicted prominently, increasing image resolution, or improving the model, e.g., by exploring the use of vision transformers \citep{yuan2021}. Updating the model would not affect the pipeline in any way: The detection models are inputs to the pipeline and can be replaced with any preferred model. While the differentiation between straight pipes, elbows and T-fittings worked well in the example test environments, adding them as different object classes in the 'Object detection and and instance prediction' module could make the results more robust as it would make the use of crude heuristics redundant. The models could also be extended to detect more objects than pipes, pumps, tanks, and valves. This would pave the way for applying the pipeline to environments other than hydraulic systems. The latter would at the same time help to validate the method further. While the pipeline shows promising results on generated scenes that mimic real infrastructure, it remains a prototype until it can be validated on a real-world example. Thus, the main objective for the future remains testing and validating the pipeline on an actual critical infrastructure.


\section{Conclusion}\label{sec:conclusion}

This article proposes a prototypical graph generation pipeline that shows the relations between relevant predetermined objects that are instrumental to the type of critical infrastructure in question. For example, this study investigates hydraulic systems for which pipes, reducers, expanders, tanks, valves and pumps were identified as relevant. The pipeline is based on a combination of photogrammetry, deep learning for object detection and instance segmentation, and heuristics for inferring relations between objects. The use of these methods has the advantages of being cost-efficient (both hardware for data collection and computation) and accessible. The user-defined set of rules for the 'Graph generation' module makes it easy to tailor the pipeline to specific use cases and transfer it from one problem to the next, while its transparency and explainability are vital for the high stakes decision-making required for critical infrastructure.

\singlespacing


\section*{Acknowledgments}
We would like to thank our colleagues Sebastian Sporrer, Lena Schreiber and Norman Wilhelms for their generous assistance, the valuable discussions and their helpful feedback. We further would like to extend our gratitude to our colleagues Marius Stürmer and Kai Franke for the conceptualization and implementation of the test environments in the Unreal Engine. This research would not have been possible without their meticulous work.

\section*{Author Contributions}
MD: Conceptualization, Formal Analysis, Investigation, Methodology, Visualization, Writing - original draft, Writing - review and editing. YT: Conceptualization, Data curation, Formal Analysis, Writing - original draft, Writing - review and editing.

\section*{Conflict of Interest Statement}
The author(s) declared that this work was conducted in the absence of any commercial or financial relationships that could be construed as a potential conflict of interest.

\section*{Funding}
The author(s) declared that financial support was received for this work and/or its publication. This research was funded by the project Automated Model Generation (AMG) within the German Aerospace Center (DLR).

\bibliographystyle{unsrtnat}
\bibliography{references}

\end{document}